\documentclass[12pt,a4paper]{article}
\usepackage[utf8]{inputenc}
\usepackage[english]{babel}
\usepackage[T1]{fontenc}
\usepackage{amsmath, amssymb, fullpage, graphics}
\usepackage{caption, graphicx, fancyhdr, fancybox}
\usepackage{color}
\usepackage{tabularx}
\usepackage{hyperref}
\usepackage{cite}

\newcommand{\bR}{\mathbb{R}}

\usepackage{pgf}
\usepackage[numbers]{natbib}
\newtheorem{definition}{Definition}[section]    

\makeatletter
\fancypagestyle{footmark}{
  \ps@@fancy 
  \fancyfoot[C]{\footmark}
}

\makeatother
\begin{document}

\begin{center}
{\large\sf Analyze the Effects of Weighting Functions on Cost Function \\in the Glove Model}
\vspace{2mm}

Trieu Hai Nguyen$^{1}$ \\[2mm]

${^1}$ Faculty of Information Technology, Nha Trang University, \\
02 Nguyen Dinh Chieu Street, Nha Trang City, Vietnam \\[2mm] 
e-mails: trieunh@ntu.edu.vn

\end{center}
\begin{abstract}
When dealing with the large vocabulary size and corpus size, the run-time for training Glove model is long, it can even be up to several dozen hours for data, which is approximately 500MB in size. As a result, finding and selecting the optimal parameters for the weighting function create many difficulties for weak hardware. Of course to get the best results, we need to test benchmarks many times. In order to solve this problem, we derive a weighting function, which can save time for choosing parameters and making benchmarks. It also allows one to obtain nearly similar accuracy at the same given time without concern for experimentation.

\end{abstract}


\section{Introduction}
Representing language is the key to machines that can communicate like humans. Thus, methods for vector representations of words like word embeddings are research trends and become more popular. In recent years, some methods learning language representation as Word2Vec, GloVe, FastText in Word Embeddings have created a great revolution in the field of Natural Language Processing \cite{mikolov, pennington, fasttext2016, fasttext2017}. In word embeddings, words or phrases will be mapped to real-value vectors, which are input feature vectors  of machine learning or deep learning models. They are commonly used in text classification task, information retrieval, question answering, semantically and syntactically related tasks. Some classical types of word embeddings as One-hot encoding, Count Vector, TF-IDF Vector based on frequency of words (tokens) in each document. These types are the simplest way to vectorize words. However, they have some disadvantages like the size of vocabulary is large but the important information it stores is not much and the semantic word similarities are not represented. The problem of semantic word similarities can be solved by using Co-occurrence matrix, but this method also waste store resource in high-dimensional feature space. 

In order to overcome these drawbacks of the previous methods, Milkolov proposed Word2Vec method, which can be applied for learning high-quality word vectors without limiting the size of data sets \cite{mikolov}.  Unlike a Co-occurrence matrix, Word2Vec is a neural network model, which has only one hidden-layer (projection layer). Word vector representations will learn from the above neural networks by mapping words to the target words. Some basically model architectures of Word2Vec are Continuous Bag-of-Words (CBOW) and Skip-gram Model. These model architectures are also known as New Log-linear Models in the way defined by Mikolov. In CBOW model, target word will be predict from the given context words and vice versa for Skip-gram Model. 

Inheriting from Word2Vec, the GloVe model, which was introduced by Pennington et al. in 2014, contains the advantages combined from the global matrix factorization based on latent semantic analysis (LSA) methods and local context window methods based on Skip-gram Model \cite{Deerwester90indexingby,pennington}. According to LSA's idea for reducing the dimensional of word representations from co-occurrence matrix,  the authors modified the type of co-occurrence matrix in the form ``term-term'', which means rows and columns are words in the vocabulary of datasets. The values of elements in the matrix correspond to the number of occurrences of whole context words for each target word in a given context window. After that, context windows will be scanned over the entire documents by using skip-gram model. In particular, this model is only concerned with the nonzero elements in the co-occurrence matrix and generates vector space with meaningful sub-structure. As a consequence, GloVe model obtains efficient statistical information and works well on the word analogy task.

Besides the outstanding advantages of the GloVe model, it also has inconvenience like weighting function in model depends heavily on empirical parameters. Moreover, the model also depends on several context windows  factors for each different dataset and language. Finding and selecting the optimal parameters for the weighting function to obtain the best results on several tasks has led to some difficulties. In particular, the most notable waste is the cost for calculation and time with large corpus. In order to drastically decreases the waste, in this work, we derive a new weighting function based on theory without having to perform many parameter tests. In section \ref{section_a}, we analyze the GloVe model proposed in \citep{pennington} and give the other formulation of the weighting function. In addition, we also prove that new weighting function fits the properties of Pennington et al. through function theory. In section \ref{section_b}, we make some comparisons between our new function with the linear ($\alpha=1$) and experimental version ($\alpha=3/4$) considered in \citep{pennington}.

\section{Analyzing GloVe model}{\label{section_a}}
First, we remind some basic steps for constructing GloVe model from the combination of the statistics of word in a corpus and semantic relationships between the word vectors. The word-word co-occurrence matrix denoted by $M$. Where $M_{i,j}$ is the number of occurrences of word $j$ in the context of word $i$.  $M_i=\sum_{k=1}^{|V|} M_{ik}$ is defined as the number of times any word occurs in the context of word $i$, which equals to the total values of the columns for the corresponding row $i$ in co-occurrence matrix, $|V|$ is the number of words in the vocabulary or also known as the size of the vocabulary. A probability formula of the target word $j$ appears in the context of word $i$ has the form
\begin{equation}
P_{ij}=P(j|i)=\frac{M_{ij}}{M_i}=\frac{M_{ij}}{\sum_{k=1}^{|V|} M_{ik}}. \label{eq:specific_probability}
\end{equation}
An example to understand the above formula is $C=S1 \cup S2$, where $S1$ and $S2$ are sentences in corpus $C$. We assume that
\[
S1=\textit{``NTU is not a small university''}, \qquad
S2=\textit{``NTU is a big university''}.
\]
This corpus generates vocabulary $V=\textit{``NTU, is, not, a, small, university, big''}$. The word-word co-occurrence matrix $M$ with the given context window is expressed as
\begin{table}[htb]
\begin{center}
\caption{Co-occurrence matrix with context window equals to \textit{1}}
\begin{tabular}{|c|c|c|c|c|c|c|c|}
\hline 
$\downharpoonleft$ target--context$\rightharpoonup$ & NTU & is & not & a& small & university & big \\ 
\hline 
NTU & 0 & 2 & 0 &0 & 0 & 0 & 0 \\ 
\hline 
is & 2 & 0 & 1 & 1 & 0 & 0 & 0 \\ 
\hline 
not & 0 & 1 & 0 & 1 & 0 & 0 & 0 \\ 
\hline 
a & 0 & 1 & 1 & 0 & 1 & 0 & 1 \\ 
\hline 
small & 0 & 0 & 0 & 1 & 0 & 1 & 0 \\ 
\hline 
university & 0 & 0 & 0 & 0 & 1 & 0 & 1 \\ 
\hline 
big & 0 & 0 & 0 & 1 & 0 & 1 & 0 \\ 
\hline 
\end{tabular} 
\end{center}
\end{table}

The probability of some target words in the context of words $\textit{``NTU''}$ and $\textit{``is''}$ are $P_{\textit{NTU}, \textit{NTU}}=0$, $P_{\textit{NTU}, \textit{is}}=1$ and $P_{\textit{is}, \textit{NTU}}=1/2$ respectively. From these probabilities, the question is that how we can extract semantic relation between words directly from the co-occurrence matrix. Following \citep{pennington}, we can take the relation of meaning by using the ratio of their co-occurrence probabilities. In order to show the power of co-occurrence matrix, let's back to a similar example in \citep{pennington}. We give an example, which represents traffic signals rules, here $i$, $j$ are \textit{``go''} and \textit{``stop''} respectively. Similarly, we define a variable called $z$, which represents a few words related to $i$, $j$. The relationship of these words are shown in table \ref{table:traffic_signals}. Usually, \textit{``green''} light corresponds to the word \textit{``go''} and did not relate to the word \textit{``stop''}, which leads to the conclusion that the ratio $P(z|i)/P(z|j)$ should be large. However, in case $z = fashion$ did not relate to the word \textit{``go''} or \textit{``stop''}. Thus the ratio of these probabilities will be closed to 1. Thanks to these ratios, we can distinguish separate words (\textit{green} and \textit{red}) from irrelevant words (\textit{yellow} and \textit{fashion}).
\begin{table}[htb]
\begin{center}
\caption{Co-occurrence probabilities of words in traffic signals example} \label{table:traffic_signals}
\begin{tabular}{|l|c|c|c|c|}
\hline 
Probability \& Ratio & $z=green$ & $z=red$ & $z=yellow$ & $z=fashion$\\ 
\hline 
$P(z|i=go)$ & high & low & high & low\\ 
\hline 
$P(z|j=stop)$ & low & high & high & low\\ 
\hline 
$P(z|i)/P(z|j)$ & $>1$ & $<1$ & $\approx1$ &$\approx1$ \\ 
\hline 
\end{tabular} 
\end{center}
\end{table}

Based on the ratios of co-occurrence probabilities, which can be used as arguments of word vector learning, the general model converts these ratios into word vectors written in the form
\begin{equation}
F(w_i,w_j,\tilde{w}_z)=P_{iz}/P_{jz}, \label{eq:general_model}
\end{equation}
where probabilities $P$ were calculated from the formulas \eqref{eq:specific_probability}, $F$ is an arbitrary function, $\{w, \tilde{w}\}\in\mathbb{R}^d$ are the word vectors of three words $i$, $j$ and $z$. Especially, symbol $\tilde{w}$ denotes separate context word vectors. Obviously, the equation above has some difficulties like too many $F$ functions satisfy equation \eqref{eq:general_model}, there are three input arguments in the $F$ function, the ratio value in the right-hand side (RHS) is scalar while the left-hand side (LHS) is  vectors.  
For the first difficulty, we can restrict the number of functions $F$ based on the analogy between words belongs to vectors $w$. The analogous can be computed through the definition of linear vector space
\begin{definition} \label{def:linear_space}
    Let $S=\{w_i,w_j\}$ be a vector space over the field $K$ on which is defined an operation of addition and an operation of multiplication by scalars, for all scalars $c \in K$. These operations must satisfy some following conditions
\begin{enumerate}
\item vector addition is commutative law: $w_i+w_j=w_j+w_i$, for all vectors $w_i$, $w_j$ in $S$,
\item scalar multiplication is distributive law: $c(w_i + w_j)=cw_i+cw_j$, for all $c$ in $K$ and $w_i$, $w_j$ in $S$.
\end{enumerate}     
\end{definition}
Considering the definition \ref{def:linear_space}, we use subtraction of $w_i$ and $w_j$ to find the difference between these vectors. The equation \eqref{eq:general_model} simplifies to
\begin{equation}
F(w_i-w_j,\tilde{w}_z)=P_{iz}/P_{jz}. \label{eq:subtraction}
\end{equation}

\begin{figure}[htb]
\vspace*{-3mm}
\begin{center}
\resizebox{0.6\textwidth}{!}{\includegraphics{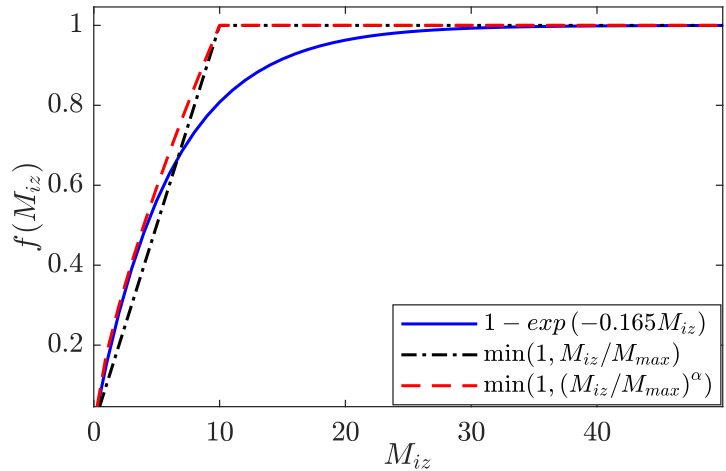}}\\[0pt]
{\caption{The solid line is our weighting function, dashed-dotted line and dashed line are original weighting function with $\alpha=1$ and $\alpha=3/4$ respectively. Parameter $M_{max}=10$ was used for both dashed cases.}\label{fig:parameters}} 
\end{center}
\end{figure}

Currently, LHS only has two vector arguments, in order to overcome the challenge of the number of arguments, taking into account the minimization of input arguments to the LHS is needed. We can use the dot product of vectors to convert LHS into scalar in RHS. Thus the LHS of \eqref{eq:subtraction} can be rewritten as
\begin{equation}
F\left( (w_i-w_j)^T \tilde{w}_z \right)=F\left( w_i^T\tilde{w}_z-w_j^T\tilde{w}_z \right), \label{eq:dot_product}
\end{equation}
where symbol $T$ refers to the transposition matrix to be compatible with the dimensions in the dot product. Looking back to the first drawback, the specific function $F$ can be found by assuming that $F$ is homomorphism between the groups $G=(\bR , -)$ and $H=(\bR^+ , /)$
\begin{definition} \label{def:homomorphism}
Let $G=(\bR , -)$ and $H=(\bR^+ , /)$ be groups. A homomorphism $F$: $G \longmapsto H$ is a function $F$: $G \longmapsto H$ such that, for all $g_1$, $g_2$ $\in G$, 
\[F(g_1-g_2)=F(g1)/F(g_2).\]
\end{definition}
Applying definition \eqref{def:homomorphism} to equation \eqref{eq:dot_product} gives
\[
F\left( w_i^T\tilde{w}_z - w_j^T\tilde{w}_z \right)=F\left( w_i^T\tilde{w}_z\right)\left/F\left(w_j^T\tilde{w}_z \right)\right.,
\]
which leads to an expression
\begin{equation}
\frac{F\left( w_i^T\tilde{w}_z\right)}{F\left(w_j^T\tilde{w}_z \right)}=\frac{P_{iz}}{P_{jz}}, \qquad 
F\left( w_i^T\tilde{w}_z\right)=cP_{iz}, \label{eq:homomorphism}
\end{equation}
where constant $c=F\left(w_j^T\tilde{w}_z \right)/P_{jz}$. We assume that this constant does not change the form of our relationship and can be neglected. Under this assumption, equation \eqref{eq:homomorphism} can be combined with the probability formula at equation \eqref{eq:specific_probability} to transform into
\begin{equation}
F\left( w_i^T\tilde{w}_z \right)=M_{iz}/M_i. \label{eq:neglected_c}
\end{equation}
Following the homomorphism theorems \citep{halbeisen}, the mapping 
\begin{align*}
F:& G \longmapsto H\\
& x \longmapsto e^x
\end{align*}
is an isomorphism, and for all $x \in G$: $F^{-1}=\text{ln}$. Using this property as the solution of function $F$ in LHS \eqref{eq:neglected_c}, which implies $F\left( w_i^T\tilde{w}_z \right)=e^{w_i^T\tilde{w}_z}$. Substituting this result into equation \eqref{eq:neglected_c}, yields the relation
\[w_i^T\tilde{w}_z=\text{ln}\left( M_{iz}/M_i \right),\]
which is equivalent to
\begin{equation}
w_i^T\tilde{w}_z=\text{ln}M_{iz} - \text{ln}M_{i} \label{eq:ln_subtraction}
\end{equation}
In RHS of the above equation, we can observe that only term $\text{ln}M_{iz}$ is related depending on the word $z$. Hence the term $\text{ln}M_{i}$ will be eliminated. However, after eliminating this term, it is necessary to keep the symmetry of equation \eqref{eq:ln_subtraction} by adding bias terms of the network $b_{w_i}$, $b_{w_z}$ for $w_i^T$ and $\tilde{w}_z$ respectively. We can rewrite the equation \eqref{eq:ln_subtraction} in the form
\begin{equation}
w_i^T\tilde{w}_z+b_{w_i}+b_{w_z}-\text{ln}M_{iz}=0 \label{eq:core_of_GloVe}
\end{equation}
Clearly, the above equation is easier and simpler to write a cost function than equation  \eqref{eq:general_model}. However, it still faces a few issues such as $\text{ln}M_{iz}$ is undefined when $M_{iz}=0$, the weight of all elements in co-occurrence matrix is the same while some rare words are noisy or carry little information. Normally, we can use laplacian smoothing $\text{ln}M_{iz}\rightarrow \text{ln}(M_{iz}+1)$ to fix the logarithm divergences. The second issue can be solved by redistributing weights for elements. Specifically, the authors of GloVe model proposed a new weighting function $f(M_{iz})$ in cost function to overcome all  issues. After combining least mean square function for \eqref{eq:core_of_GloVe} with the new definition of weighting function above, yields the cost function in the form
\begin{equation}
J=\sum\limits_{\{i,z\}=1}^{|V|} f(M_{iz})\left( w_i^T\tilde{w}_z+b_{w_i}+b_{w_z}-\text{ln}M_{iz} \right)^2, \label{eq:cost_function}
\end{equation} 
where $i$, $z$ can be considered as target word and context word respectively, $f$ must satisfy some of the properties mentioned in \citep{pennington} and it takes the form  
\begin{equation}
f(x)=min\left( 1, \left(x/x_{max} \right)^{\alpha} \right), \label{eq:weighting_function}
\end{equation}
in which $x_{max}$ and $\alpha$ are the empirical parameters. In \citep{pennington}, they found that for $\alpha = 3/4$ and $x_{max}=100$, the result was the best when compared to the baseline of a linear version $\alpha=1$. However, equation \eqref{eq:weighting_function} is overly dependent on empirical parameters, which leads to many difficulties for weak hardware in finding and selecting the optimal parameters for the weighting function. The foregoing causes a waste of time and resources in the benchmark performing. Thus, we derive a new weighting function, which reduces the dependence on finding and selecting parameters from the experiment. Our new weighting function is expressed as
\begin{equation}
g(x)=1-e^{-0.165x}.\label{eq:our_weighting_function}
\end{equation}
Fig.\ref{fig:parameters} represents some weighting functions. The solid line describes our function, which is proposed in \eqref{eq:our_weighting_function}. The dashed-dotted and dashed lines show Pennington's weighting function with parameters used are $\alpha=1$ (linear version) and $\alpha=3/4$ (optimal version) respectively. $M_{max}=10$ corresponds to $x_{max}$ in function \eqref{eq:weighting_function}. In addition, our function still satisfies the three properties of the Pennington's weighting function such as $g(0)=0$, $g(x)$ is non-decreasing and  $g(x)$ is relatively small for large value of $x$. The advantages of the function $g(x)$ over $f(x)$ are
\textit{
\begin{enumerate}
\item $\lim\limits_{x \rightarrow \infty}g(x)=1$  without limiting the condition $x <x_{max}$ as in function $f(x)$,
\item greatly reducing time and resources in the benchmark to find and optimize empirical parameter $\alpha$.
\end{enumerate}
}

\section{The comparison between models} {\label{section_b}}
In this section, we apply our new weighting function to GloVe model and compare results with the original version in \citep{pennington}. The dataset used in this test is ``text8''\footnote{The dataset is available at \url{http://mattmahoney.net/dc/textdata.html}}, which is the shortening of Wikipedia Text and has 100MB in size. text8 includes five types of semantic questions and nine types of syntactic questions. When processing data, this set creates a vocabulary of size \textit{71290}, \textit{253854} unique words and 17005207 tokens. Now, we use the GloVe model on this dataset to compare the quality of different weighting functions. First, we evaluate the similarities among words in the semantic-syntactic word relationship. In order to measure the similarity between  words we use cosine distance \citep{mikolov}. Table \ref{table:similar} shows the semantic and syntactic relationships of word pairs through the cosine distance in two cases, which corresponds to our weighting function $g(x)$ and Pennington's weighting function $f(x)$. According to results in the table \ref{table:similar}, we can observe semantic similarities between countries like \textit{VietNam--Laos}, \textit{VietNam--China} and \textit{VietNam--Cambodia} as well as syntactic relationships like \textit{big--biggest} and \textit{small-smaller}.
\begin{table}[htb]
\begin{center}
\caption{Some similarities between words at iteration \textit{15}}\label{table:similar}
\begin{tabular}{|c|c|c|c|}
\hline 
\multicolumn{2}{|c|}{Word Pair} & Cosine distance--$g(x)$  & Cosine distance--$f(x)$ \\ 
\hline 
vietnam & laos & 0.621690 & 0.689430 \\ 
  & china & 0.615755 & 0.608853 \\ 
  & cambodia & 0.607484 & 0.642829 \\ 
\hline 
russia & ukraine & 0.780213 & 0.785777 \\ 
  & germany & 0.771556 & 0.777929 \\ 
  & romania & 0.747980 & 0.727222 \\ 
\hline 
usd & dollars & 0.762856 & 0.766871 \\ 
\hline 
work & works & 0.840226 & 0.841410 \\ 
\hline 
big & biggest & 0.541842 & 0.492286 \\ 
\hline 
small & smaller & 0.843994 & 0.848305 \\ 
\hline 
brother & sister & 0.784377 & 0.711776 \\ 
\hline 
\end{tabular} 
\end{center}
\end{table}

\begin{figure}[!htb]
\begin{center}
\resizebox{1\textwidth}{!}{\includegraphics{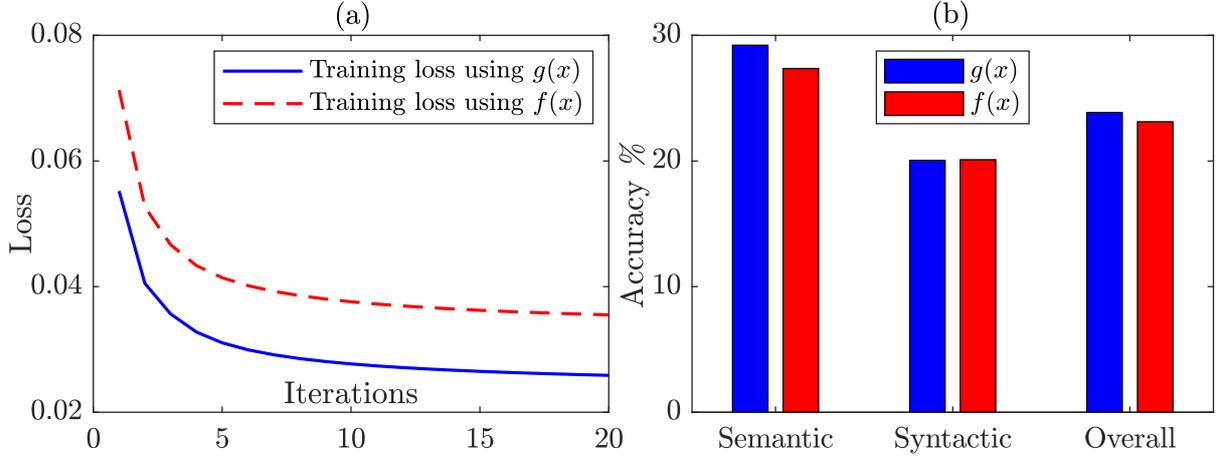}}\\[0pt]
{\caption{(a) Training loss at iteration \textit{20}. Solid line, dashed line describe the values of cost function $J$ in equation \eqref{eq:cost_function} using weighting functions $g(x)$ and $f(x)$ respectively. (b) The accuracy of GloVe model on the analogy task for all question types at iteration \textit{20}. The first and the second group are the accuracies on subsets of the semantic and syntactic respectively. The third group is a comparison of the overall results of two weighting function versions. The blue and red columns correspond to the accuracy using functions $g(x)$ and $f(x)$.
}\label{fig:loss_accuracy}} 
\end{center}
\end{figure}

Furthermore, we also provide an overview of the accuracy of the GloVe model using function $g(x)$ and $f(x)$ at iteration \textit{20}. The parameters used in the GloVe model are $vector\_size=50$, $context\_window=15$, $x\_max=10$ and $\alpha=3/4$. Training loss of GloVe model with two weighting function versions at iteration \textit{20} is shown in Fig.\ref{fig:loss_accuracy}a, where, the solid line illustrates the result of the cost function using $g(x)$ and the dashed line corresponding to $f(x)$ is used. In Fig.\ref{fig:loss_accuracy}a, we can easily observe that the $J$ value when using $g(x)$ will converge to zero faster than $g(x)$. Besides Fig.\ref{fig:loss_accuracy}a, we also performed the comparison of accuracy on the analogy task when using $g(x)$ and $f(x)$ in Fig.\ref{fig:loss_accuracy}b. The results in this figure, which leads to the conclusion that when using GloVe model with our new weighting function will give slightly better results than the original weighting function version in \citep{pennington}. Specifically, function $g(x)$ gives the general result of \textit{23.86}\% and \textit{23.12}\% for function $f(x)$.
\begin{figure}[!htb]
\begin{center}
\resizebox{0.55\textwidth}{!}{\includegraphics{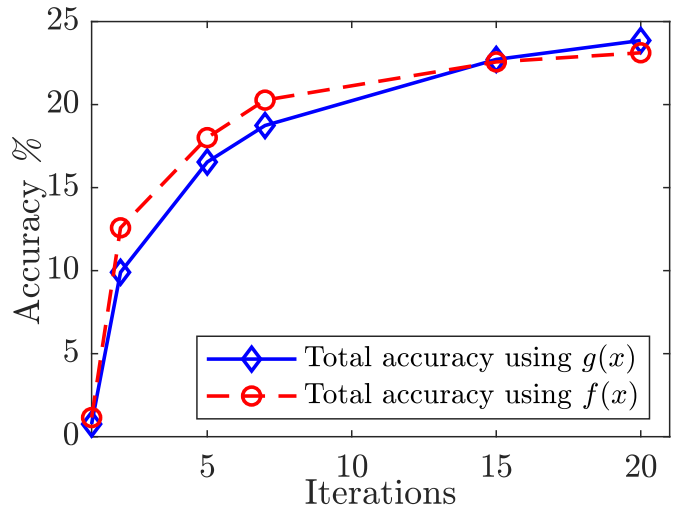}}\\[0pt]
{\caption{Total accuracy of GloVe model using $g(x)$ and $f(x)$
}\label{fig:accuracy_iteration}}
\end{center}
\end{figure}

Moreover, in Fig.\ref{fig:accuracy_iteration}, we also show the results through each iteration. The solid line with diamond markers corresponds to total accuracy using our weighting function. The dashed line with circle markers represents total accuracy using the original function. We noticed that from iteration \textit{15} our results start a little better than the original version.

\section{Conclusion}
In this work, we researched and modified the weighting function in the cost function of GloVe model proposed in \citep{pennington}. The construction of our weighting function is entirely based on the properties of a family of functions given by Pennington et al.(2014). Our new weighting function reduces the dependence of GloVe model on empirical parameters more than the original version. Through the new function found in this work, 
it helps us to save time and resources for choosing parameters and making benchmarks on weak hardware. In particular, the results obtained are nearly similar accuracy at the same given time without concern for experimentation when compared to the original version. Further, from iteration \textit{15}, our results are a little better on the same dataset.  

\bibliography{ref}
\bibliographystyle{IEEEtranN}

\end{document}